# Incorporating Domain Knowledge in Deep Neural Networks for Discrete Choice Models

Shadi Haj Yahia*, Omar Mansour, Tomer Toledo

Faculty of Civil and Environmental Engineering,
Technion – Israel Institute of Technology, Haifa 32000, Israel
Emails:  shadi8@campus.technion.ac.il
omar@technion.ac.il
toledo@technion.ac.il

## ABSTRACT

Discrete choice models (DCM) are widely employed in travel demand analysis as a powerful theoretical econometric framework for understanding and predicting choice behaviors. DCMs are formed as random utility models (RUM), with their key advantage of interpretability. However, a core requirement for the estimation of these models is a priori specification of the associated utility functions, making them sensitive to modelers' subjective beliefs. Recently, machine learning (ML) approaches have emerged as a promising avenue for learning unobserved non-linear relationships in DCMs. However, ML models are considered "black box" and may not correspond with expected relationships.

This paper proposes a framework that expands the potential of data-driven approaches for DCM by supporting the development of interpretable models that incorporate domain knowledge and prior beliefs through constraints. The proposed framework includes pseudo data samples that represent required relationships and a loss function that measures their fulfillment, along with observed data, for model training. The developed framework aims to improve model interpretability by combining ML's specification flexibility with econometrics and interpretable behavioral analysis. A case study demonstrates the potential of this framework for discrete choice analysis.

**Keywords:** Deep neural networks, discrete choice models, domain knowledge, interpretability

## 1. INTRODUCTION

Discrete choice models (DCMs) have been employed widely in travel demand analysis and prediction to understand individuals' decision-making processes (Ben-Akiva et al., 1985). Most DCMs are formed as random utility models (RUMs), which assume that individuals make decisions based on utility maximization decision protocols (McFadden, 1974).

Researchers have proposed various RUM-based models, incorporating different assumptions about error structures (Ben-Akiva, 1973; McFadden, 1978; Revelt & Train, 1998; Train, 2009). However, the selection of variables and their functional forms in the utility function is subjective and can significantly impact the estimation results. Utility functions describe how different attributes of each alternative are valued by individuals, which varies depending on personal characteristics and preferences that differ across decision-makers. As such, these relationships are often nonlinear, and selecting a suitable utility model that accurately captures these complex relationships through functional forms and variable transformations can be a challenging task.

Several studies have shown that an incorrectly assumed utility functional form (e.g., linear) can cause bias in parameter estimates and in the resulting predictions. This modeling uncertainty has been a persistent concern for modelers (Torres et al., 2011).

Recently, the use of data-driven approaches to learn DCM specifications has emerged as a promising avenue to overcome the limitations of RUM specifications. Machine learning (ML) methods (e.g., neural networks, decision trees, ensemble learning) can learn non-linear mapping functions (Bishop, 2006). Deep neural networks (DNN) (LeCun et al., 2015) are an increasingly used data-driven approach that has shown higher prediction accuracy in many tasks. DNN architectures represent the decision-making process by imitating the structure of neuron activity patterns and memory. Unlike RUM, DNN models require essentially no a priori knowledge about the nature of the true underlying relationships among variables and choices. However, the main focus of their use has been on prediction at the level of the individual rather at the level of market shares.

In DNN models, input variables are fed into the network and pass through multiple hidden layers before reaching the output layer. The output layer has the same dimension as the number of alternatives, with each output representing the score $y_j$ of each alternative. These scores are then transformed into choice probabilities using a softmax function in the form $e^{y_k}/\sum_{j \in C} e^{y_j}$. While DNNs do not have the same behavioral constraints as RUMs and no utility functions to be defined, the similarity of DNN to logit models promotes their use as a more flexible substitute to logit models and interpret these scores as utilities. The main focus of these works has been on prediction accuracy (Chang et al., 2019; Mahajan et al., 2020; van Cranenburgh & Alwosheel, 2019). They find that DNNs can outperform RUMs in prediction accuracy (Hagenauer & Helbich, 2017; Hillel et al., 2019; Omrani, 2015).

However, deep neural networks (DNNs) are often criticized for their "black box" nature, which hinders the interpretation of the extracted relationships and limits their usefulness in providing insights into the factors that influence choice behavior. Unlike RUMs, DNN models do not provide explicit information about how the input variables affect the output, which may not be consistent with domain knowledge. For example, the signs of the effects of various variables on the choices (e.g., in travel choices, negative own-elasticity of choices to their costs and travel times) and ranges of marginal rates of substitutions among variables (e.g., values of time) may not align with the expected relationships. While the high flexibility of DNNs allows them to find complex non-linear specifications without a priori knowledge, their ability to explain and interpret the choice behavior and generalize the model to unseen situations are essential features, which may be negatively affected by the "black box" form of these models (Van Cranenburgh et al., 2021).

Works have been done to promote the use of ML models for choice modeling by "opening" the "black box" through the development of external tools to derive economic behavioral information that enhance their interpretability. Wang, et al., (2020) numerically extract different economic quantities of interest from DNNs, including estimates of individual choice predictions and probabilities, market shares, social welfare, substitution patterns of alternatives, probability derivatives, elasticities, and heterogenous marginal rates of substitution. They show that most economic information, when averaged over several trained models, is reasonable and more flexible than multinomial logit (MNL) models. However, at the disaggregate level, some results are counter intuitive, such as negative values of time.

Another technique to estimate choice sensitivities relies on partial dependence plots (PDPs) (Friedman, 2001). It calculates choice probabilities for every possible value of a variable for each observation. Zhao et al., (2020) implement and visualize PDPs to capture both magnitude and direction of effects. The insights gained from are qualitative in nature. While these plots are computationally intuitive and are easy to implement, they provide a clear interpretation only when features are uncorrelated. The assumption behind this method is that the features are independent, while this is often not the case in choice modeling context. If this assumption is violated, the averages calculated for the PDP will include data points that are very unlikely or even impossible to exist. In their case study they compare several ML models to MNL. In some of cases, the PDPs for ML models deviate substantially from linearity, which suggests that ML models can capture nonlinear relationships among the explanatory variables and choice.

A second approach involves combining ML and RUMs. Sifringer et al., (2018, 2020) introduced a systematic utility function that incorporates both an interpretable and a learned part. The interpretable part is specified explicitly by the modeler, as is common with RUMs, while the learned part utilizes a DNN to discover additional useful utility terms from available data. The DNN is fed with explanatory variables and outputs a term that captures additional utility to alternatives. The model improved the estimation log-likelihood value compared to a traditional MNL. However, a potential limitation of this approach is that the DNN term is unbounded, and its relationship with the explanatory variables remains unknown, which may cause problems when the model is applied with new inputs. The decision on which variables to include in each part of the utility function is also subjective and made by the modeler. Therefore, it would be beneficial to bound the DNN terms and develop a more objective way to determine which variables enter each part of the utility function.

A third approach is to use ML models with utility-like architectures to resemble RUM. Wang et al., (2020) develop an alternative specific utility deep neural network (ASU-DNN) model. Their model maintains separate utility functions for each alternative, which are trained using separate neural networks using only the variables relevant to that alternative. Thus, the utility scores for each alternative depend only on its own attributes. Systematic heterogeneity is captured by inclusion of socio-demographic variables that first enter a separate DNN and then integrated into each of the utility functions to obtain the final outputs. This model was more readily interpreted compared to fully connected DNNs and achieved comparable or even better fit to the training and testing data. But because of the model structure, as in MNL, the model exhibits the independence of irrelevant alternatives property. In addition, the model architecture might still suffer from unreasonable relationships among explanatory variables and choices.

The above-mentioned methods focus on interpretability but do not specify and enforce any restrictions on the relationships among the explanatory variables and choices (Alwosheel et al., 2019, 2021). Therefore, they are unable to guarantee that the resulting non-linear relationships are controllable and consistent with domain knowledge. Inconsistency with domain knowledge also limits model's application in prediction for evaluating new policies in new scenarios. Prediction for new policy analysis might be needed beyond the fitting region. In such cases, extrapolation is required, which is a challenge for ML models. Therefore, the ability to explain and interpret the choice behavior to understand the factors that affect it and generalize the ML model to unseen situations are essential features. By incorporating domain knowledge into the model, extrapolation can be made possible in prediction.

This study aims to enhance the consistency of ML model with domain knowledge. The proposed framework allows incorporating domain knowledge by the introduction of constraints to the model. This is facilitated by generating pseudo data that contains such knowledge to be fed to the model, in combination of formulating a loss function to penalize violating this knowledge. The contribution of this work is to demonstrate that domain knowledge can be incorporated while preserving flexibility. The proposed approach is independent of the model structure, making it possible to implement on other architectures. It demonstrates that theory and expert knowledge can be introduced effectively and flexibly as required by domain expertise and will promote the application of data-driven ML models for travel choice predictions through building trust and giving modelers control over the model's behavior.

The rest of the paper is organized as follows: the next section describes the proposed methodology for incorporating domain knowledge in DNN. The following section presents a case study that implements the proposed framework and discusses the results of different models. Conclusions and potential enhancements and extensions to the proposed methodology are presented in the final section.

## 2. METHODOLOGY

The idea behind incorporating domain knowledge in DNNs involves augmenting the data given to the model and modifying the loss function that the model optimizes. To achieve this, additional data, termed pseudo data, is generated to hold the targeted knowledge that the model is expected to capture. The loss function is then formulated to include terms that use this data, in combination with the original model loss function, such as the negative log-likelihood. The additional loss terms measure the extent to which the trained model is consistent with the domain knowledge.

The overall framework for incorporating domain knowledge into DNNs is shown in Figure 1. The framework is independent of the model structure, allowing seamless integration with existing off-the-shelf DNN architectures. The model is trained on two sets of inputs: the originally available observed data and domain knowledge, which is mathematically formulated as a set of constraints on the outcomes of the trained model. The observed data represents the available dataset collected, including socio-economic characteristics of decision makers, attributes of the alternatives, and choices. The domain knowledge represents the knowledge that the modeler wants to incorporate into the model and expects to be captured (e.g., directions of sensitivities).

In this work, the modeler's prior expectations are related to directions of the effects of an alternative's attributes on its own utility. For example, these may be negative effects of mode travel times and costs on the utilities of these modes. In this case, the model is constrained to learn a monotonically decreasing probability of choosing an alternative with respect to its travel time and cost and, consequently, monotonically increasing probabilities of choosing the remaining alternatives.

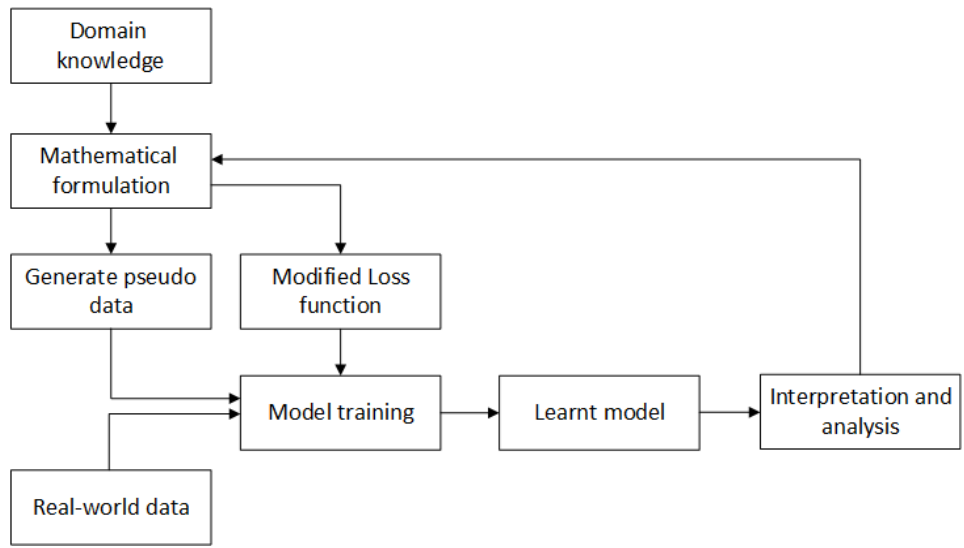

Figure 1. Overall framework for incorporating domain knowledge

Consider a training set consists of $N$ samples $\{(x_i, y_i)\}$, $i = 1, \ldots, N$, where $x_i$ is a feature vector in $x \in \mathbb{R}^{\mathcal{D}}$, and $y_i$ is the discrete choice among $\mathcal{C}$ alternatives, $y_i \in \{1, \ldots, \mathcal{C}\}$. Let $p_c(x_i)$ be the probability of choosing alternative $c$ given input $x_i$, and $x_i[m]$ is the value of feature $m$ in the feature vector. The estimated model is considered to be monotonically increasing in $p_c$ with respect to feature $m$ if $p_c(x_i) \geq p_c(x_j)$ for any two feature vectors $x_i, x_j$, such that $x_i[m] \geq x_j[m]$ and $x_i[h] = x_j[h]$, for all $h \in \mathcal{D}\setminus m$. The opposite applies for decreasing monotonicity. The rest of the components are described as follows.

### *Generating Pseudo Data*

Following monotonicity constraints above, pseudo data can be generated as pairs of samples to numerically approximate the probabilities' derivatives that are constrained. For each monotonicity constraint with respect to a feature $m$, $K$ pseudo samples are generated uniformly along the region values of that feature $x_{k,1}^*$. Each pseudo sample is then paired with another pseudo one, such that the second pseudo sample has a positive incremental change applied to feature $m$. The relationship required for an increasing monotonicity constraint of probability of choosing alternative $c$ with respect to feature $m$ is $p_c(x_{k,2}^*) - p_c(x_{k,1}^*) \geq 0$.

The pseudo data does not require labels (i.e., chosen alternatives), as they are only used for capturing domain knowledge, not for predicting the chosen alternative. This ability to generate pseudo samples enhances the model in three ways:

1. When the dataset is small, the pseudo dataset helps increase the dataset size to learn the model's parameters.
2. When the input feature region is imperfectly covered, the pseudo data helps fill gaps and enforce the model to learn along the full range of possible values.
3. Generating pseudo data outside the range of current values for specific features helps enforcing better learning, hence enabling extrapolation in the outer regions (i.e., unseen scenarios).

### *Loss Function*

The loss function includes two components: prediction loss and domain knowledge loss. The prediction loss quantifies the accuracy of predictions and can be calculated for example using the negative log-likelihood ($\mathcal{L}_{NLL}$) method commonly used in RUMs. This calculation is performed only for samples with observed choices and is represented by the following formula:

$$\mathcal{L}_{NLL} = - \sum_{i=1}^{N} \sum_{c \in \mathcal{C}} g_{i,c} \cdot \log(p_{i,c}) \qquad (1)$$

Where $g_{i,c}$ equals 1 if alternative $c$ is chosen by individual $i$ and 0 otherwise.

The domain knowledge loss measures the violation of monotonicity constraints on the probability of choosing alternative $c$ with respect to feature $m$. This is determined using pseudo sample pairs that estimate the derivatives of the probabilities, represented by the following formula:

$$\mathcal{L}_{c,m} = \sum_{k=1}^{K} \max\left(0, d_{c,m} \cdot \frac{p_c(x_{k,2}^*) - p_c(x_{k,1}^*)}{\Delta x_m^*}\right) \qquad (2)$$

Where $d_{c,m}$ equals 1 if the probability of choosing alternative $c$ with respect to feature $m$ should be increasing and -1 otherwise.

If it is assumed that when the probability of choosing alternative $c$ with respect to feature $m$ is in one direction, the probability of choosing other alternatives should be in the opposite direction, the total loss to be minimized can be expressed as follows:

$$\min \mathcal{L}_{total} = \mathcal{L}_{NLL} + \sum_{m \in M} \sum_{c \in C} w_{c,m} \cdot \mathcal{L}_{c,m} \quad (3)$$

Where $M$ represents the indices of the features that constrain the probabilities, and $w_{c,m}$ represents the weight of each constraint violation penalty.

## Model Training

The training process is illustrated in Figure 2. Observed data, represented as vector **x**, and a vector of pseudo sample pairs $\mathbf{x}^* = \{(x^*_{1,1}, x^*_{1,2}), \ldots, (x^*_{k,1}, x^*_{k,1})\}$ are fed into the model. The total loss, calculated as a combination of the prediction loss from the observed samples **x** and the domain knowledge loss from the pseudo data $\mathbf{x}^*$, is minimized using the backpropagation technique. This process continues iteratively until convergence is achieved.

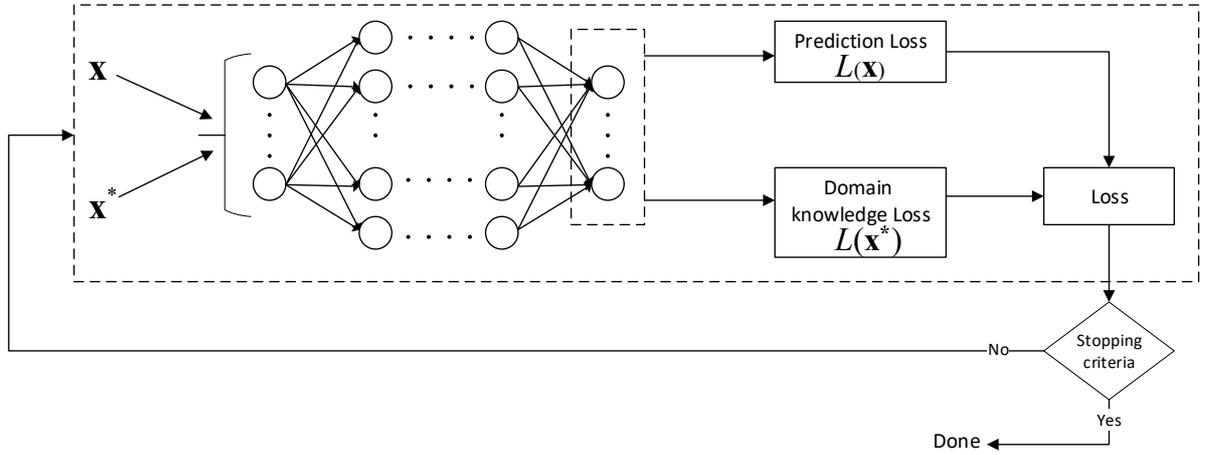

Figure 2. Model training process

## 3. CASE STUDY

The methodology outlined above was applied to a mode choice dataset to assess the potential of incorporating domain knowledge in a DNN model and examine the impact of such knowledge on the resulting economic information.

## Dataset

The experiment relies on the openly available Swissmetro dataset (Bierlaire et al. (2001)). The Swissmetro dataset is a stated preference survey that was collected in Switzerland in 1998. Participants were asked to provide information regarding their preferred mode of transportation between the new Swissmetro (SM) mode, car, and train. Each individual reported the chosen transport mode for various trips among the three alternative modes. Variables used from the dataset are described in **Table 1**. Observations with unavailable alternatives, unknown features or outlier values were filtered, resulting in 7,778 samples. The dataset was then divided into training, validation, and testing sets in the ratio of 60:20:20.

Table 1. Swissmetro dataset variables descriptions

| Variable | Descriptions |
|---|---|
| Train travel time | Train travel time [minutes] |
| Train cost | Train cost [CHF] |
| Train headway | Train headway [minutes] |
| SM travel time | SM travel time [minutes] |
| SM cost | SM cost [CHF] |
| SM headway | SM headway [minutes] |
| Car travel time | Car travel time [minutes] |
| Car cost | Car cost [CHF] |
| Seats | Seats configuration in the Swissmetro (dummy). Airline seats (1) or not (0). |
| Group | Different groups in the population. 2: current rail users, 3: current road users |
| Purpose | Travel purpose. 1: Commuter, 2: Shopping, 3: Business, 4: Leisure, 5: Return from work, 6: Return from shopping, 7: Return from business, 8: Return from leisure |
| First | First class traveler (0 = no, 1 = yes) |
| Luggage | 0: none, 1: one piece, 3: several pieces |
| Age | It captures the age class of individuals. The age-class coding scheme is of the type: 1: age≤24, 2: 24<age≤39, 3: 39<age≤54, 4: 54<age≤ 65, 5: 65 <age |
| Male | Traveler's Gender 0: female, 1: male |
| Income | Traveler's income per year [thousand CHF] 0 or 1: under 50, 2: between 50 and 100, 3: over 100 |
| TRAIN AV | Train availability dummy |
| CAR AV | Car availability dummy |
| SMAV | SM availability dummy |

*Experimental Design*

The proposed methodology was implemented on two model architectures: DNN and ASU-DNN. The DNN model was an off-the-shelf model, while the ASU-DNN model was proposed by Wang et al., (2020a) and calculates alternative-specific utilities. Both models were estimated in both an unconstrained and a constrained (i.e., with domain knowledge) version. The constrained models are referred to as C-DNN and C-ASU-DNN, respectively. In addition, a Multinomial Logit (MNL) model was also estimated for comparison.

The domain knowledge incorporated in the constrained models includes negative own-sensitivities of choice probability to travel time and cost and positive cross-sensitivities. All constraints are incorporated simultaneously, and are mathematically formulated as follows:

$$\frac{\partial P_j}{\partial travel\ time_j} \leq 0, \forall j \quad (4)$$

$$\frac{\partial P_j}{\partial cost_j} \leq 0, \forall j \quad (5)$$

$$\frac{\partial P_i}{\partial travel\ time_j} \geq 0, \forall i \neq j \quad (6)$$

$$\frac{\partial P_i}{\partial cost_j} \geq 0, \forall i \neq j \tag{7}$$

Where $P_j$, $travel\ time_j$, and $cost_j$ are the choice probability, travel time and cost of alternative $j$. Therefore, in total 18 constraints are incorporated into the model.

To evaluate the performance of each model, negative log-likelihood and prediction accuracy were measured on each of the datasets. In addition, predicted market shares were calculated. To further illustrate the fulfillment of domain knowledge, choice probabilities were presented. Finally, to analyze the effect of domain knowledge on the extracted economic information, values of time (VOT) were calculated from all models.

## 4. RESULTS

*Prediction performance*

Table 2 presents the negative log-likelihood (NLL) and accuracy (Acc.) of each estimated model in the training, validation, and testing sets. The results demonstrate that the DNN model achieves the best NLL and accuracy across all sets, indicating its strong capability for empirical fitting to data. However, its average NLL and accuracy show a significant decrease in testing, indicating its vulnerability to overfitting.

The C-DNN model shows a significant drop in fit compared to DNN in training, but slightly on testing. However, its performance is similar on training and testing sets, which indicates that incorporating domain knowledge may improve model's generalizability. ASU-DNN has a more restricted structure than DNN, leading to a lower performance. C-ASU-DNN is the least flexible model among the four DNNs, and therefore it performs the worst on testing. MNL performance is the weakest in both training and testing sets because of its simple linear specification. It is worth noting, however, that the maximum drop in performance between constrained and unconstrained versions of DNN and ASU-DNN models on testing is limited to 0.02 points in average NLL and 1.6% in accuracy.

Table 2. Average negative log-likelihood (NLL) and prediction accuracy

| Model | Training | | Validation | | Testing | |
|---|---|---|---|---|---|---|
| | Avg. NLL | Acc [%] | Avg. NLL | Acc [%] | Avg. NLL | Acc [%] |
| DNN | 0.51 | 78.3 | 0.58 | 75.5 | 0.68 | 70.1 |
| C-DNN | 0.70 | 69.1 | 0.66 | 71.9 | 0.70 | 69.1 |
| ASU-DNN | 0.61 | 74.0 | 0.67 | 69.0 | 0.72 | 69.4 |
| C-ASU-DNN | 0.62 | 74.2 | 0.67 | 68.8 | 0.73 | 67.8 |
| MNL | 0.73 | 67.6 | 0.70 | 71.0 | 0.77 | 66.1 |

*Market shares*

While prediction accuracy relates to predicting choices at the level of individuals, transportation policy planners are mainly interested in prediction at the market level. Table 3 shows the predicted market shares by the different models and the root mean square error (RMSE) in each model. In the training set, unconstrained models DNN and ASU-DNN outperform their constrained counterparts due to their high flexibility in empirical fitting to data. However, in the testing set, constrained models show better performance than unconstrained ones, highlighting the significance of domain knowledge in improving generalizability on unseen data. Furthermore, although MNL yields exact market shares in estimation, it demonstrates the poorest performance when tested on unseen data. These findings emphasize the trade-off between model flexibility and generalizability and highlight the importance of incorporating domain knowledge for accurate market-level predictions.

Table 3. Market shares of travel modes

| | **Training set** | | | | | |
|---|---|---|---|---|---|---|
| | DNN | C-DNN | ASU-DNN | C-ASU-DNN | MNL | Observed |
| Train | 6.4% | 6.6% | 6.8% | 5.8% | 6.8% | 6.8% |
| SM | 56.4% | 57.8% | 55.9% | 56.8% | 56.0% | 56.0% |
| Car | 37.2% | 35.6% | 37.4% | 37.3% | 37.3% | 37.3% |
| RMSE | 0.3% | 1.4% | 0.1% | 0.7% | 0% | |

| | **Validation set** | | | | | |
|---|---|---|---|---|---|---|
| | DNN | C-DNN | ASU-DNN | C-ASU-DNN | MNL | Observed |
| Train | 6.5% | 6.2% | 6.4% | 5.5% | 6.9% | 4.5% |
| SM | 58.1% | 58.3% | 58.0% | 59.2% | 58.1% | 59.6% |
| Car | 35.4% | 35.5% | 35.7% | 35.3% | 35.1% | 35.9% |
| RMSE | 1.5% | 1.3% | 1.5% | 0.7% | 1.7% | |

| | **Testing set** | | | | | |
|---|---|---|---|---|---|---|
| | DNN | C-DNN | ASU-DNN | C-ASU-DNN | MNL | Observed |
| Train | 6.6% | 6.7% | 6.4% | 5.5% | 6.7% | 8.1% |
| SM | 54.9% | 55.6% | 54.6% | 55.9% | 53.5% | 54.7% |
| Car | 38.5% | 37.7% | 39.0% | 38.7% | 39.8% | 37.2% |
| RMSE | 3.4% | 2.9% | 3.6% | 2.8% | 4.4% | |

*Choice probabilities*

To assess whether the models fulfill expected domain knowledge, choice probability functions were calculated for each of the six variables for the three alternatives. Each variable's value was systematically varied across all observations by a percentage ranging from -50% to +50%, while the remaining variables kept unchanged. These plots illustrate how the model's behavior responds to changes in variable values (e.g., for evaluating a new policies) on an aggregated level.

The estimated coefficients in the MNL are with the expected sign (i.e., negative coefficients of travel time and cost in all utility functions), therefore, the directions of choice probabilities are consistent with domain knowledge as can be seen in Figure 3-4.

However, choice probabilities may not always be consistent with domain knowledge when derived from the unconstrained models, even in ASU-DNN where utilities are calculated independently from the others following RUM. This inconsistency could be restrained when domain knowledge is incorporated into the models.

For example, the expected behavior of the estimated models is that train choice probabilities decrease as its travel time increases, while SM and car choice probabilities increase. However, the DNN model presented in Figure 3(a) shows an unusual pattern where SM choice probability increases as train travel times decrease by up to 20%, but decreases when train travel times are increased by up to 25%. Moreover, as train travel times increase by more than 20%, car choice probabilities decrease, making SM more appealing to travelers. Figure 3(c) shows the predicted choice probabilities provided by the ASU-DNN model. The model behaves consistently until train travel times increase by 30%, where car choice probabilities slightly decrease, while car choice probabilities slightly increase. These inconsistencies are addressed in the constrained models C-DNN and C-ASU-DNN in Figure 3(b) and Figure **3**(d), respectively.

Similar behaviors are expected as a function of changes in train cost. However, the DNN model presented in Figure 4(a), shows that car choice probability increases as train travel costs decrease by up to 30%, and it decreases when train costs are increased up to 10%. These results suggest that, while flexible, the DNN model performs poorly in unseen scenarios beyond the fitting region, even at an aggregated level. Conversely, these inconsistencies are avoided in C-DNN (Figure 4(b)). ASU-DNN presents slight inconsistencies when train costs increase by more than 40%, where train and car choice probabilities flip directions (Figure 4(c)). This inconsistency is mitigated in the C-ASU-DNN model as shown in Figure 4(d).

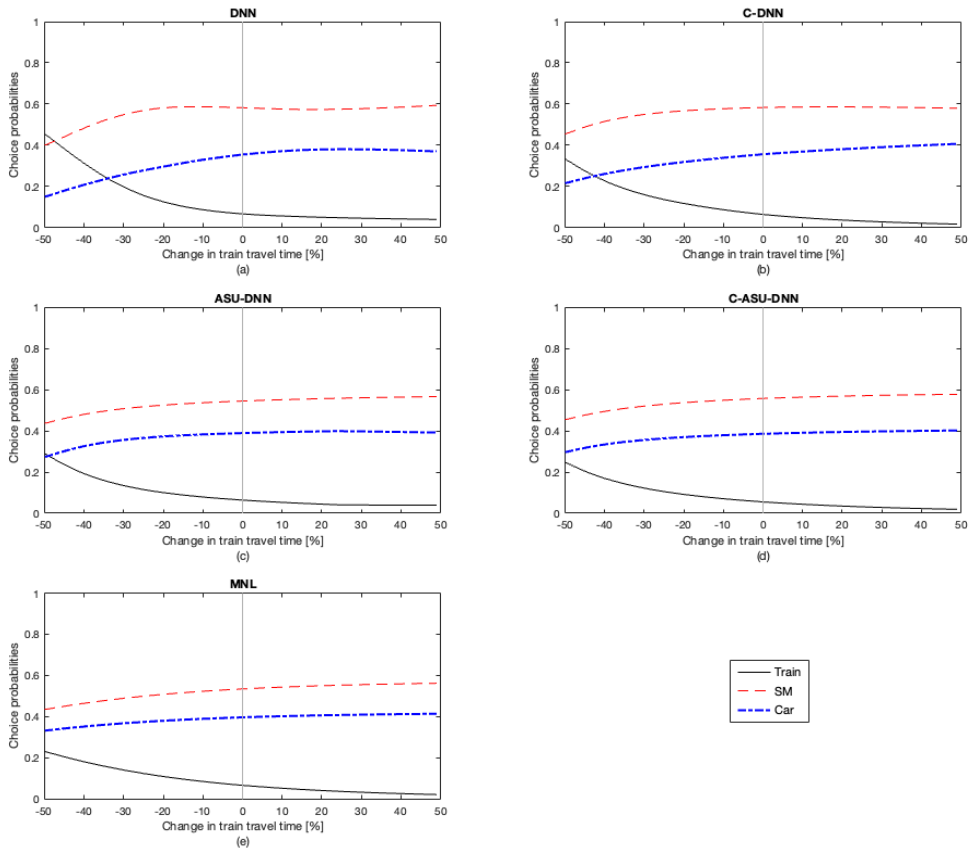

Figure 3. Alternatives' choice probabilities as a function of a percentage change in train travel time

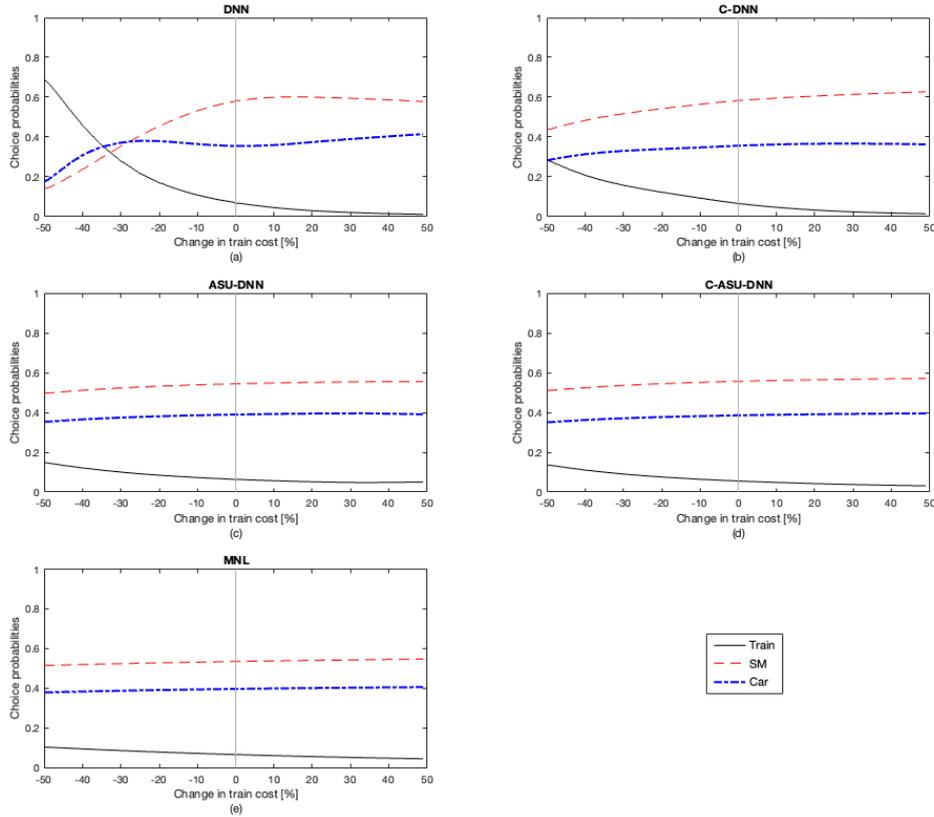

Figure 4. Alternatives' choice probabilities as a function of a percentage change in train cost

*Value of time*

Value of time (VOT) is an important economic information that is obtained from DCMs which represents travelers' willingness to pay in order to save time and is used to evaluate the benefits of transport projects. VOT for each alternative mode was calculated given each of the five models. In MNL, VOT is a single value obtained by taking the ratio of travel time and cost coefficients for each alternative. Table 4 presents the descriptive statistics of the calculated VOTs, including mean, median, and the percentages of negative VOTs.

In the analysis of the calculated VOTs, it was found that the unconstrained models yield mean VOT values that appear reasonable for each alternative, but they obscure more complex results. Specifically, significant percentages of VOTs in these models have negative values, which contradicts common domain knowledge. On the other hand, the constrained models, C-DNN and C-ASU-DNN, which incorporate domain knowledge through the modeling process, offer more reliable results. In these models, VOT is calculated as the ratio of probability derivatives with respect to travel time and cost, which are mostly constrained to be positive, resulting in a significant decrease in the percentages of negative VOTs for all alternatives.

Examining the median VOTs in Table 4, it is evident that the DNN model tends to underestimate mean VOT for all modes compared to other models, potentially due to the substantial number of negative VOTs. In contrast, MNL tends to overestimate the mean VOT, which could result from biased parameter estimates due to linear utilities misspecification.

Table 4. Value of time descriptive statistics: mean, median, and percentage of negative VOT

| Value of Time (Train) | Mean [CHF/h] | Median [CHF/h] | Negative |
|---|---|---|---|
| DNN | 11.9 | 13.9 | 24.7% |
| C-DNN | 26.2 | 24.9 | 0% |
| ASU-DNN | 45.6 | 65.3 | 4.0% |
| C-ASU-DNN | 54.1 | 63.1 | 1.5% |
| MNL | 101.0 | 101.0 | 0% |
| Value of Time (SM) | Mean [CHF/h] | Median [CHF/h] | Negative |
| DNN | 10.6 | 15.3 | 31.2% |
| C-DNN | 61.7 | 50.9 | 0.1% |
| ASU-DNN | 105.6 | 41.9 | 6.1% |
| C-ASU-DNN | 355.0 | 47.1 | 1.0% |
| MNL | 83.4 | 83.4 | 0% |
| Value of Time (Car) | Mean [CHF/h] | Median [CHF/h] | Negative |
| DNN | 90.0 | 37.3 | 28.0% |
| C-DNN | 83.6 | 79.4 | 0% |
| ASU-DNN | -84.6 | 99.6 | 5.4% |
| C-ASU-DNN | 2746.6 | 107.0 | 0.6% |
| MNL | 200.2 | 200.2 | 0% |

Figure 5-7 present the distribution of heterogenous VOT in the sample from all models for train, SM, and car, respectively. VOTs derived from DNN model, seem to be normally distributed for all modes (Figure 5(a), Figure 6(a) and Figure 7(a)). In contrast, the C-DNN model manages to almost completely prevent negative VOTs. The derived VOTs for all modes are log-normal distributed.

The analysis of VOT distributions for ASU-DNN and C-ASU-DNN also reveals differences between the two models for the different modes. In the case of train (Figure 5), both models provide close median VOTs, with C-ASU-DNN showing a larger mean due to fewer negative values. For SM, both models exhibit bimodal VOT distributions (Figure 6 (c-d)), with the median values being close, but the mean from C-ASU-DNN being three times larger than that from ASU-DNN. This can be attributed to the larger number of negative values in ASU-DNN and the presence of outliers in C-ASU-DNN on the extreme positive side. For car (Figure 7), ASU-DNN yields a log-normal distribution of VOTs, while C-ASU-DNN provides a normal distribution. However, the extreme large values from C-ASU-DNN lead to an unrealistically large mean VOT, while extreme negative values from ASU-DNN result in a negative mean. Notably, ASU-DNN exhibits larger gaps between the mean and median VOTs for all modes due to the presence of extreme values. While domain knowledge helps to reduce the number of negative VOTs, it does not address the issue of extreme values, as shown in C-ASU-DNN.

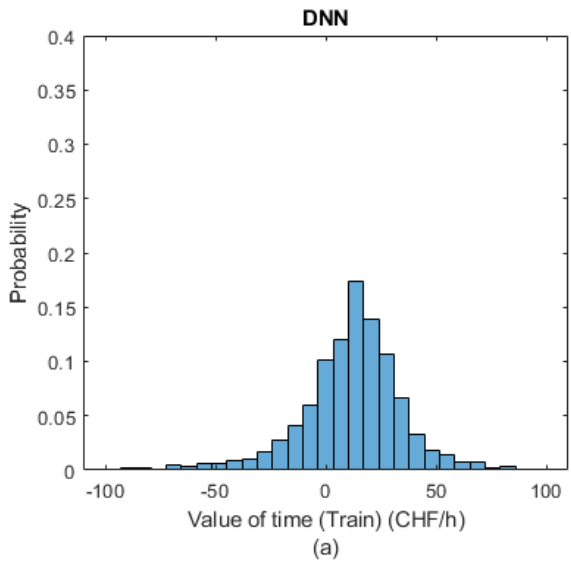
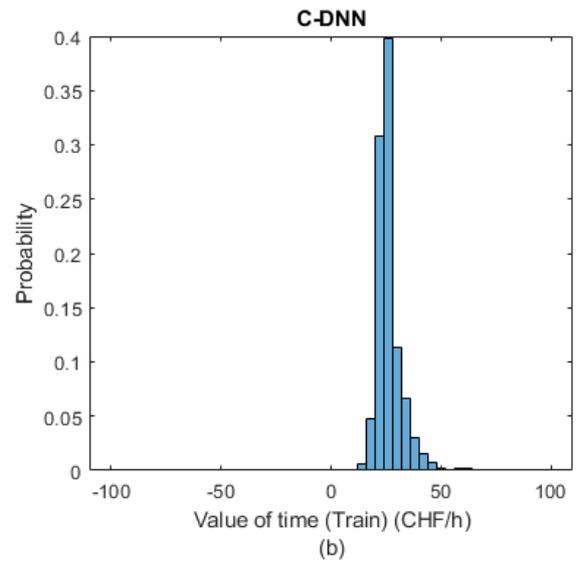
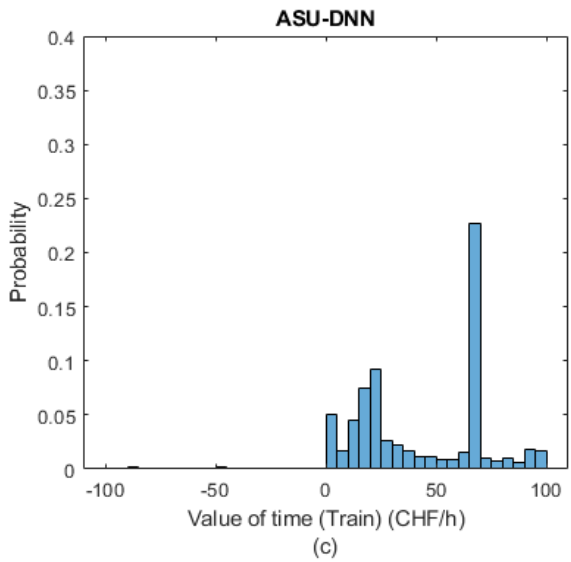
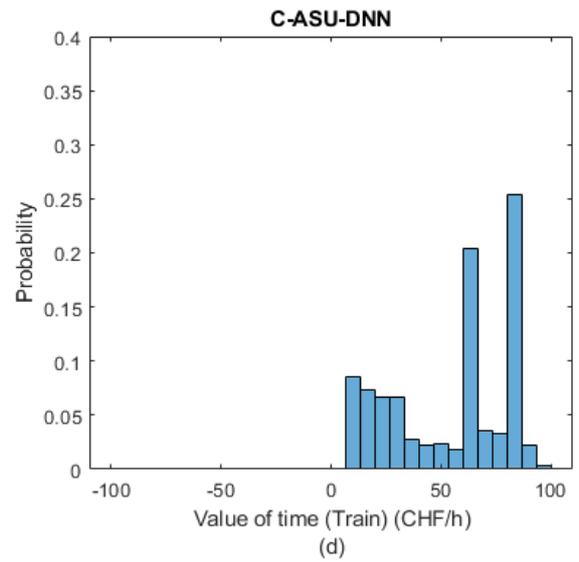

Figure 5. Heterogeneous values of time for train; the extremely large and small values are cut off from histogram

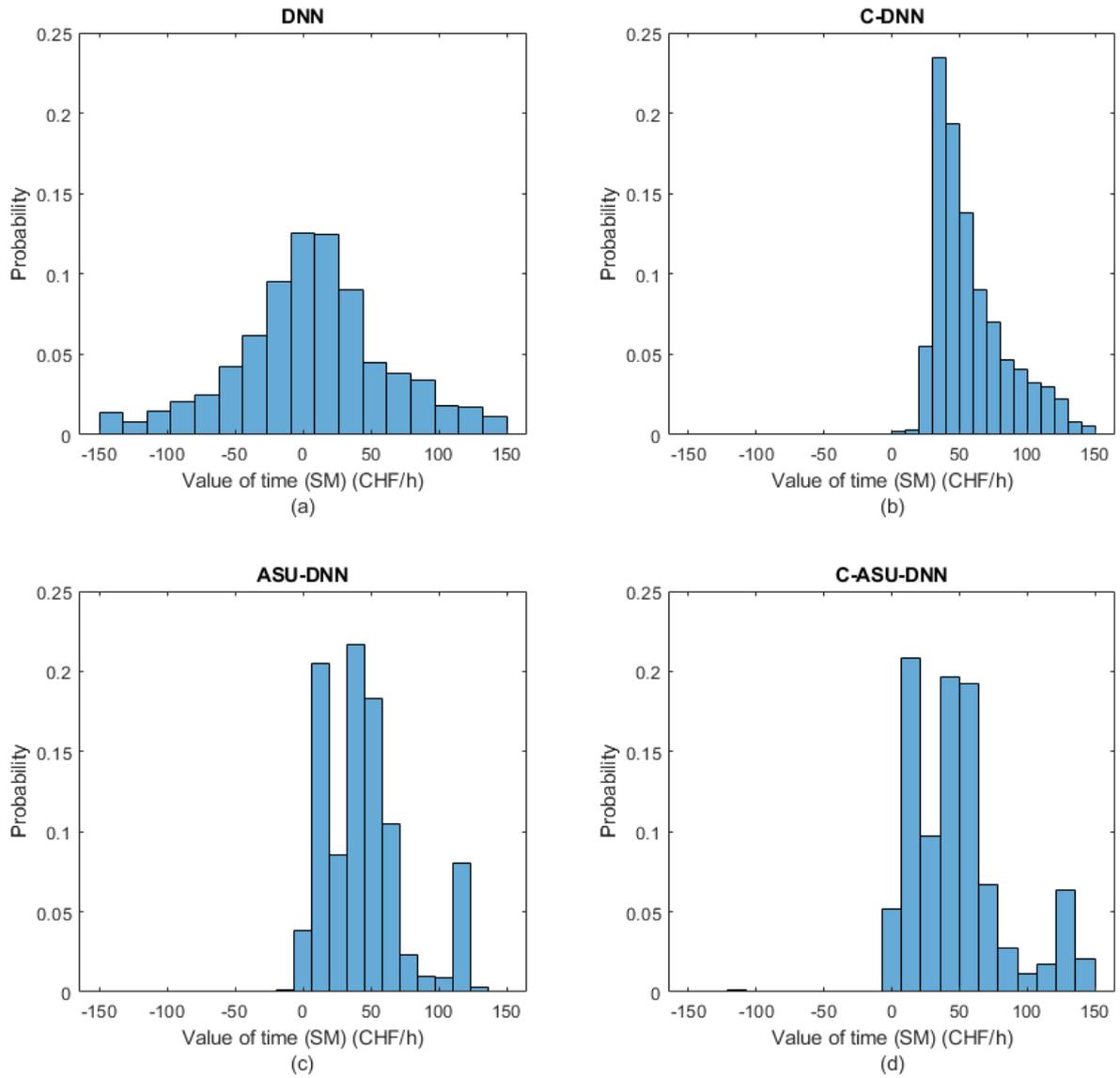

Figure 6. Heterogeneous values of time for SM; the extremely large and small values are cut off from histogram

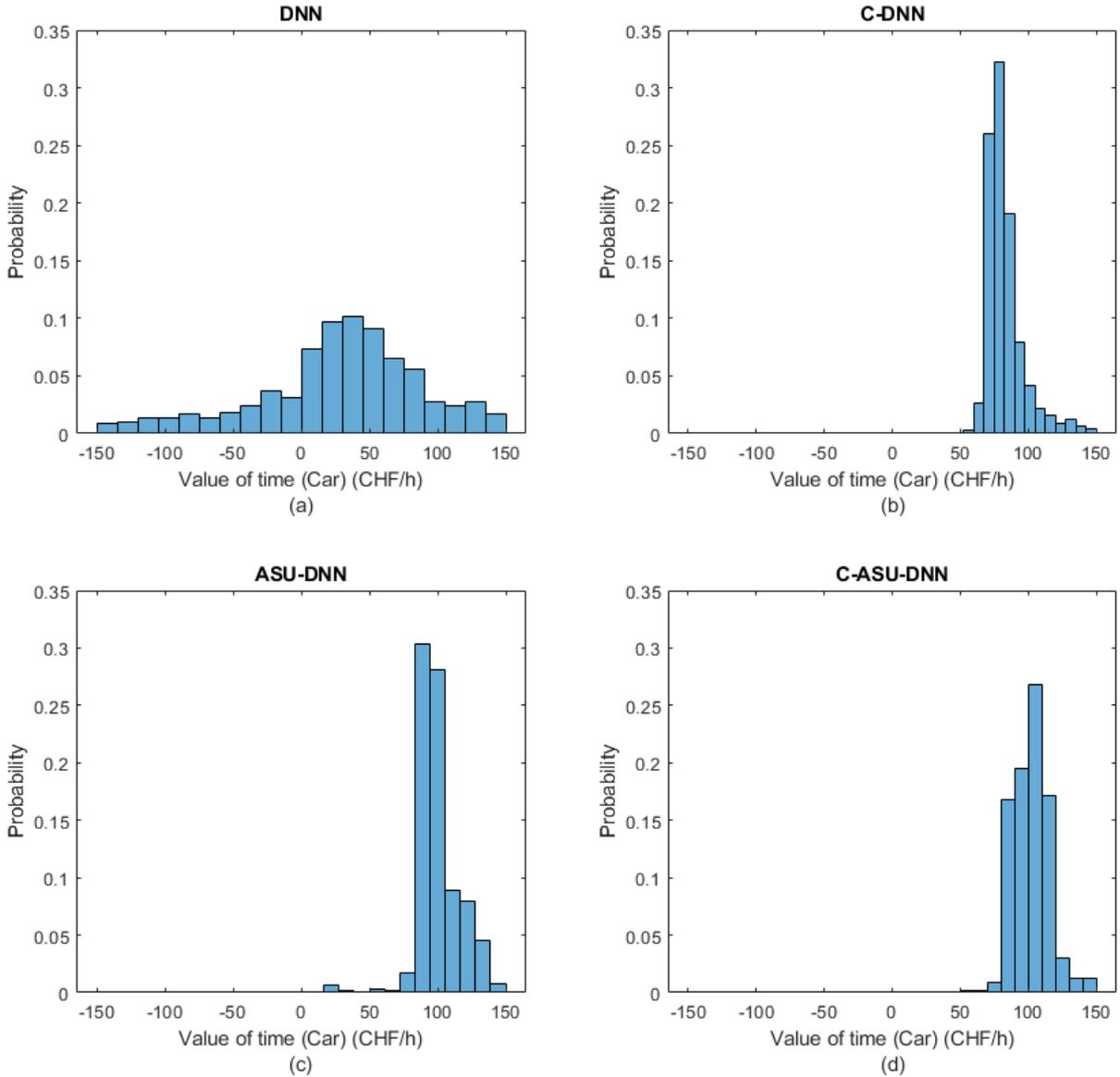

Figure 7. Heterogeneous values of time for car; the extremely large and small values are cut off from histogram

## 5. CONCLUSIONS

This study addresses a major limitation in the application of DNN models for discrete choice analysis, which often leads to counterintuitive results due to the lack of domain knowledge. Analysts typically possess knowledge that should be captured by the model, such as the negative impact of travel time on an alternative's choice probability. However, as the case study demonstrates, such knowledge is not always captured by data-driven models that rely solely on the data.

To overcome this limitation, a framework was proposed to enhance the consistency of ML models with domain knowledge. This approach involves incorporating constraints to the model to ensure that specific relationships are fulfilled, while leaving others unrestricted for data-driven learning. Only the direction of the relationships (i.e., positive or negative) is constrained, preserving flexibility in the model's form. The proposed framework is independent of the model structure, making it easy to implement on different architectures (e.g., DNN and ASU-DNN).

The proposed methodology was applied to Swissmetro dataset using fully connected DNN and ASU-DNN models. The tradeoff between accuracy and interpretability was demonstrated, as both models' prediction performance was negatively affected, but their interpretation became more reasonable and consistent with prior expectations. The case study showcases the potential of combining domain knowledge from utility theory with DNN models for more interpretable choice analysis, regardless of the architecture. It also highlights the range of differences in analyses (e.g., predicted values of time and choice probabilities) that can arise from different models. Further work is needed to include additional domain knowledge (e.g., magnitudes of elasticity and values of time) and test it on richer datasets to explore the implications of incorporating domain knowledge for choice analysis.